\documentclass[letterpaper, 10 pt, conference]{ieeeconf}  

\IEEEoverridecommandlockouts                              

\title{\LARGE \bf Multi-Agent Inverse Reinforcement Learning in Real World Unstructured Pedestrian Crowds}

\author{Rohan Chandra$^1$, Haresh Karnan$^2$, Negar Mehr$^3$, Peter Stone$^{2,4}$, Joydeep Biswas$^2$\\
\small{$^1$University of Virginia, $^3$University of California, Berkeley, $^2$The University of Texas at Austin, $^4$Sony AI}\\
\href{https://mairl-uva.github.io/}{\small Code and Data at \textbf{mairl-uva.github.io}}}

\usepackage[colorlinks=true, citecolor=magenta]{hyperref}

\usepackage{color}
\usepackage[dvipsnames]{xcolor}
\usepackage{colortbl}
\usepackage{xcolor}
\usepackage{mathrsfs}
\usepackage{tikz}
\usetikzlibrary{positioning}
\usepackage{xspace}
\usepackage{multicol,multirow}

\usepackage{amsthm}
\usepackage[outline]{contour}
\usepackage{soul}
\usepackage[]{mdframed}
\usepackage{threeparttable}
\usepackage{pifont}
\usepackage{wrapfig}
\usepackage{amsmath, amsfonts, amssymb}   
\usepackage{mathtools}
\usepackage{pifont}%
\let\labelindent\relax
\usepackage{enumitem}
\usepackage{booktabs}
\usepackage{float}
\usepackage[linesnumbered,ruled,vlined]{algorithm2e}
\usepackage{bm}
\usepackage{subcaption}
\usepackage{makecell}
\usepackage[font=small]{caption}

\newtheorem{problem}{Problem}

\usepackage{footmisc}
\usepackage{flushend}
\usepackage[authormarkup=none]{changes}
\setaddedmarkup{{\color{blue!75!black}#1}}
\setdeletedmarkup{\protect{\color{blue!25!gray}\sout{#1}}}
\definechangesauthor[name={Chris}]{CH}

\begin{document}
\maketitle
\thispagestyle{plain}
\pagestyle{plain}
\begin{abstract}

Social robot navigation in crowded public spaces such as university campuses, restaurants, grocery stores, and hospitals, is an increasingly important area of research. One of the core strategies for achieving this goal is to understand humans' intent--underlying psychological factors that govern their motion--by learning how humans assign rewards to their actions, typically via inverse reinforcement learning (IRL). Despite significant progress in IRL, learning reward functions of multiple agents simultaneously in dense unstructured pedestrian crowds has remained intractable due to the nature of the tightly coupled social interactions that occur in these scenarios \textit{e.g.} passing, intersections, swerving, weaving, etc. In this paper, we present a new multi-agent maximum entropy inverse reinforcement learning algorithm for real world unstructured pedestrian crowds. Key to our approach is a simple, but effective, mathematical trick which we name the so-called ``\textit{tractability-rationality trade-off}'' trick that achieves tractability at the cost of a slight reduction in accuracy. We compare our approach to the classical single-agent MaxEnt IRL as well as state-of-the-art trajectory prediction methods on several datasets including the ETH, UCY, SCAND, JRDB, and a new dataset, called Speedway, collected at a busy intersection on a University campus focusing on dense, complex agent interactions. Our key findings show that, on the dense Speedway dataset, our approach ranks $1^\textrm{st}$ among top $7$ baselines with $> 2\times$ improvement over single-agent IRL, and is competitive with state-of-the-art large transformer-based encoder-decoder models on sparser datasets such as ETH/UCY (ranks $3^\textrm{rd}$ among top $7$ baselines).

\end{abstract}
\section{Introduction}
\label{sec: introduction}

Robot navigation in densely populated human environments has garnered significant attention, especially in scenarios involving tightly coupled social interactions among pedestrians, scooters, and vehicles. These scenarios~\cite{chandra2024deadlock} occur in both outdoor settings~\cite{chandra2022game, chandra2022gameplan, suriyarachchi2022gameopt, mavrogiannis2022b} and indoor spaces such as restaurants, grocery stores, hospitals, and university campuses~\cite{scand, raj2023targeted, sprague2023socialgym, chandra2023socialmapf, chandra2023decentralized, smg}. Effective navigation in such dense and cluttered human environments is challenging and requires robots to be socially compliant which involves predicting and responding to human intentions~\cite{poddar2023crowd, francis2023principles}. Many approaches have approached intent prediction \textit{implicitly} via trajectory forecasting~\cite{gupta2018social, sadeghian2019sophie, Kosaraju2019BIGAT, giuliari2020transformer, marchetti2024smemo, sun2021three, ngiam2021scene, yuan2021agentformer, salzmann2023robots, salzmann2020trajectron++, he2021where}. While trajectory forecasting has yielded impressive results on sparse crowd datasets such as ETH, UCY, and JRDB, and structured traffic datasets such as the WAYMO Motion Forecasting Dataset~\cite{waymo}, it is unclear whether they perform similarly well in more complex settings that include passing, weaving, yielding to others, etc. 


Alternatively, some approaches have modeled intent \textit{explicitly} by computing humans' reward functions via inverse reinforcement learning. Recently, Gonon and Billard~\cite{singleirl} used maximum entropy IRL (inverse reinforcement learning)~\cite{maxentirl} for achieving socially compliant navigation in pedestrian crowds. However, inherently a single-agent approach, maximum entropy IRL (MaxEnt IRL) assumes a single objective function for all the agents, which works reasonably well in sparse crowds that lack many tightly coupled interactions, as shown in~\cite{singleirl}, but fails to accommodate humans' objectives in denser and more unstructured pedestrian crowds, as we show in our experiments in Section~\ref{sec: experiments}. This limitation leads us to question: \textit{Can MaxEnt IRL be adapted to learn effective robot navigation policies for real-world multi-agent unstructured pedestrian crowds?}


We address this question using a multi-agent IRL approach, for which numerous solutions already exist in the dynamic game theory and reinforcement learning literature~\cite{natarajan2010multi, lin2019multi, yu2019multi, song2018multi, schwarting2019social, lecleach2021lucidgames, rothfuss2017inverse,kopf2017inverse, mairl}. Key to making these methods work is the critical assumption of exactly known dynamics/system models of all the agents in the environment, which, however, rarely holds in practice. To relax this assumption, one may consider approximating the dynamics with, say, a simpler model (\textit{e.g.} constant velocity or constant acceleration models). The downside of such an approximation is that the optimization-based planner does not converge. 


\noindent\textbf{Main contribution:} In this work, we show that multi-agent MaxEnt IRL provides accurate predictions of pedestrians’ motion in real world unstructured crowds, with only few training trajectories. Our key insight is a simple, but effective, mathematical trick which we name the so-called ``\textit{tractability-rationality trade-off}'' trick that relaxes the assumption of known system dynamics \ul{tractably} at the cost of a slight reduction in accuracy. Bounded rationality (assumption of agents exhibiting some randomness in their decision-making) is typically, characterized by the covariance matrix of the probabilistic policy. At a high level, the trick uses matrix conditioning (a textbook technique in numerical linear algebra) to improve the condition number of the covariance matrix, thus ensuring tractability, but at the cost of reducing uncertainty, thereby reducing bounded rationality. Notably, this trade-off not only makes multi-agent IRL tractable in unstructured settings but also paves the way for new research directions—such as computing exact bounds for this trade-off.

We use off-the-shelf dynamic game theory solvers to roll out trajectories that maximize the learned reward functions. We evaluate this new approach using a dataset collected at a busy intersection on a University campus comprising of pedestrians, scooters, bicycles, particularly focusing on tightly coupled interactions such as when agents walk towards each other, walk past each other, squeeze past each other, etc. We compare our approach to the classical single-agent MaxEnt IRL as well as state-of-the-art trajectory prediction methods on several datasets including the ETH, UCY, JRDB, and a new dataset collected at a busy intersection on a University campus focusing on complex agent interaction. 

In the remainder of this paper, we discuss related work in Section~\ref{ref: related_work}, formulate the problem in Section~\ref{sec: problem_formulation}, present our technical approach in Section~\ref{sec: mairl_pedestrian_crowds}, discuss the experiments and results in Section~\ref{sec: experiments}, and conclude in Section~\ref{sec: conclusion}.

\section{Related Work}
\label{ref: related_work}

\subsection{Inverse Reinforcement Learning}
Kalman initially explored the inference of an agent's cost function within the context of inverse optimal control for linear quadratic systems with a linear control law~\cite{kalman}. Subsequent works~\cite{abbeel2004apprenticeship,ng2000algorithms} considered learning the agent's cost, assuming demonstrations adhered to optimality conditions. However, multiple cost functions can rationalize the expert's behavior, leading to inherent ambiguities. MaxEnt IRL~\cite{maxentirl} addresses this by incorporating the principle of maximum entropy, ensuring that the derived cost function is both unique and maximally uncertain.

\noindent\textbf{Multi-agent IRL:} While previous approaches focused on learning in the single-agent regime, multi-agent IRL has been extensively studied in various settings. 
Initial efforts focused on discrete state and action spaces~\cite{natarajan2010multi}, while later research extended to general-sum games with discrete spaces~\cite{lin2019multi}.  Adversarial machine learning techniques were employed for high-dimensional state and action spaces~\cite{yu2019multi}, and generative adversarial imitation learning was extended to the multi-agent setting~\cite{song2018multi}. Additionally, the problem has been approached as an estimation task~\cite{schwarting2019social,lecleach2021lucidgames}, and linear quadratic games were considered with known equilibrium strategies~\cite{rothfuss2017inverse,kopf2017inverse}. The maximum entropy IRL~\cite{mairl} framework was introduced to accommodate bounded rationality and noisy human demonstrations.

\subsection{Trajectory Prediction}
We focus on deep learning-based trajectory and intent prediction which dates back ten years to the seminal SocialLSTM work~\cite{sociallstm}. TraPHic and RobustTP~\cite{chandra2019robusttp, chandra2019traphic} use an LSTM-CNN framework to predict trajectories in dense and complex environments.   
    TNT~\cite{Zhao2020TNTTT} uses target prediction, motion estimation, and ranking-based trajectory selection to predict future trajectories. DESIRE~\cite{Lee2017DESIREDF} uses sample generation and trajectory ranking for trajectory prediction.
    PRECOG~\cite{precog} combines conditioned trajectory forecasting with planning objectives for agents. Yoo et al.~\cite{yoo2022gin} uses a Seq2Seq framework to encode agents' observations over neighboring agents as their social context for trajectory forecasting and decision-making. Cao et al.~\cite{cao2022leveraging} uses temporal smoothness in attention modeling for interactions and a sequential model for trajectory prediction. However, a major limitation of this class of methods is that they employ behavior cloning to try and mimic the data, without inherently capturing humans' internal reward functions. 
    
\subsection{Intent-aware Multi-agent Reinforcement Learning} 
Related to our task, intent-aware multi-agent reinforcement learning~\cite{qi2018intent} estimates an intrinsic value that represents opponents' intentions for communication~\cite{wang2021tom2c} or decision-making. Many intent inference modules are based on Theory of Mind (ToM)~\cite{rabinowitz2018machine} reasoning or social attention-related mechanisms~\cite{vemula2018social, dai2023socially}. Wu et al.~\cite{wu2023multiagent} uses ToM reasoning over opponents' reward functions from their historical behaviors in performing multi-agent inverse reinforcement learning (MAIRL). Chandra et al.~\cite{chandra2023socialmapf} uses game theory ideas to reason about other agents' incentives and help decentralized planning among strategic agents. 
However, many prior works oversimplify the intent inference and make some prior assumptions about the content of intent.

\subsection{Opponent Modeling} 
Opponent modeling~\cite{he2016opponent} has been studied in multi-agent reinforcement learning and usually deploys various inference mechanisms to understand and predict other agents' policies. Opponent modeling could be done by either estimating others' actions and safety via Gaussian Process~\cite{zhu2020multi} or by generating embeddings representing opponents' observations and actions~\cite{papoudakis2021agent}. Inferring opponents' policies helps to interpret peer agents' actions~\cite{losey2020learning} and makes agents more adaptive when encountering new partners~\cite{parekh2022rili}. Notably, many works \cite{von2017minds, ndousse2021emergent} reveal the phenomenon whereby ego agents’ policies also influence opponents’ policies. To track the dynamic variation of opponents’ strategies made by an ego agent's influence, \cite{xie2021learning, wang2022influencing} propose the latent representation to model opponents' strategies and influence based on their findings on the underlying structure in agents' strategy space. Jaques et al.~\cite{jaques2019social} provides a causal influence mechanism over opponents' actions and defines an influential reward over actions with high influence over others' policies. Kim et al.~\cite{kim2022influencing} proposes an optimization objective that accounts for the long-term impact of ego agents' behavior in opponent modeling. A considerable limitation of many current methodologies is the underlying assumption that agents continually interact with a consistent set of opponents across episodes.

\noindent\textbf{\textit{Placing our work in the literature:}} Our work directly builds on top of the multi-agent IRL literature. It is complementary to the trajectory prediction, multi-agent reinforcement learning, and opponent modeling literature, which implicitly model other agent's objectives, whereas our approach explicitly models humans' intent via reward functions.

\section{Background and Problem Formulation}
\label{sec: problem_formulation}
\begin{figure*}[h!]
    \centering
    \begin{tikzpicture}[scale=0.1, node distance=.5cm and 3cm, 
    every node/.style={draw, rounded corners, minimum width=3cm, minimum height=1.5cm, scale=0.8, align=center, text width=4.5cm, align=center}]

    \node[fill=red!20, text width=5cm, align=center] (box1) {
      \begin{minipage}{5cm}
        \centering
        \textbf{Inputs} \\ 
    \begin{tabular}{l}
        \contour{black}{$\rightarrow$} System Dynamics~\eqref{eqn: constant_vel} \\
        \contour{black}{$\rightarrow$} Initial state $x_0$ \\
        \contour{black}{$\rightarrow$} Dataset $\mathcal{\widehat D} = \{ \widehat \Gamma^{1:k}\}_N$ \\
        \contour{black}{$\rightarrow$} Feature function $\Phi(\cdot)$ \\
        \contour{black}{$\rightarrow$} Initialize $\theta^i, \Gamma^i_{\theta^i}\gets \mathcal{F}^i_{\theta^i}\left(x_0 \right)$
    \end{tabular}
      \end{minipage}
    };

    \node[fill=blue!20, text width=4.5cm, align=center] (box2) [right=1cm of box1] {
      \begin{minipage}{4.5cm}
        \centering
        \textbf{Compute Expected Features} \\ 
        $\Phi^i\left( \Gamma^i_{\theta^i}\right)$
      \end{minipage}
    };

\node[above=0.05cm of box2, draw=none,  yshift = -0.5cm] {\textit{At current time $t$}};

    \node[fill=green!20, text width=8.5cm, align=center] (box3) [right=2cm of box2] {
      \begin{minipage}{8.5cm}
        \centering
        \textbf{Compute New Weights} \\ 
        $ \theta^{i,*} = \arg\min_{\theta^i}\left \lVert \mathbb{E}_{\Gamma^i_{\theta^i}\sim\mathcal{D}_{\theta^i}}\Phi^i\left( \Gamma^i_{\theta^i}\right) - \mathbb{E}_{\widehat\Gamma^i\sim\mathcal{\widehat D}}\Phi^i\left(\widehat\Gamma^i\right) \right \rVert$\\
        \big(Equations~\eqref{eq: problem_statement_sec4} and~\eqref{eq: new_theta}\big)
      \end{minipage}
    };

    \node[fill=yellow!20, text width=8.5cm, align=center] (box4) [below=2cm of box3] {
      \begin{minipage}{8.5cm}
        \centering
       \textbf{New Policy Rollout via} \\ 
        $\mathcal{F}^i_{\theta^i}\left(x_t \right)$ \big(Equation~\eqref{eq: H}\big)\\\vspace{5pt}
        \textit{If $\left ( H^i_t\right)^{-1}$ is non-PSD, perform the Tractability-Rationality Trade-off trick (Equation~\eqref{eq: trick})}
      \end{minipage}
    };

    \draw[->, line width=0.5mm] (box1) -- (box2);
    \draw[->, line width=0.5mm] (box2) -- node[pos=0.8, yshift = -1.2cm, below, draw=none, font = \Large, sloped] {\textit{Until convergence}}(box3);
    \draw[->, line width=0.5mm] (box3) --  node[pos=0.5, xshift=-1.75cm, right , draw=none, font = \Large] {$\theta^i_\textrm{new}$}(box4);
    \draw[->, line width=0.5mm] (box4) -| node[pos=0.2, yshift = -0.3cm, above, draw=none, font = \Large, sloped] {$\Gamma^i_{\theta^i_\textrm{new}}$}(box2);

    \end{tikzpicture}
    \caption{\textbf{Flow diagram of the multi-agent inverse reinforcement learning algorithm.} The algorithm begins by taking as input the system dynamics, initial state, dataset, feature function, and initial parameters. At each time step, it computes the expected features based on the current policy. These features are used to update the weights by minimizing the difference between the expected and ground truth features. The updated weights are then used to update the game-theoretic policy and roll out a new set of trajectories. If the Hessian matrix is non-positive definite, the ``Tractability-Rationality trade-off'' trick is applied to ensure tractability in unstructured environments.}
    \label{fig: overview}
\end{figure*}

We define a game, $G \coloneqq \left \langle k, \mathcal{X}, \mathcal{T}, \{\mathcal{U}^i\}, \{\mathcal{J}^i\}\right\rangle$, where $k$ denotes the number of agents. Hereafter, $i$ will refer to the index of an agent and appear as a superscript whereas $t$ will refer to the current time-step and appear as a subscript. The general state space $\mathcal{X}$ (\textit{e.g.} SE(2), SE(3), etc.) is continuous; the $i^\textrm{th}$ agent at time $t$ has a state $x^i_t\in \mathcal{X}^i$. 
Over a finite horizon $T$, each agent starts from $x^i_{0} \in \mathcal{X}^i$ and reaches a goal state $x^i_T \in \mathcal{X}^i$.
An agent's transition function is a mapping, $\mathcal{T}^i:\mathcal{X}\times \mathcal{U}^i \longrightarrow \mathcal{X}$, where $u^i_t \in \mathcal{U}^i$ is the continuous control input for agent $i$. A discrete trajectory for agent $i$ is specified by the sequence $\Gamma^i = \left( x^i_{0},u^i_{0}, x^i_1,u^i_{1}, \ldots, x^i_{T-1}, u^i_{T-1}, x^i_T  \right)$. 

Each agent has a parameterized running cost\footnote{Note that broadly speaking, reward maximization is equivalent to cost minimization, which is what we model in this work.} $\mathcal{J}^i:\mathcal{X} \times \mathcal{{U}} \longrightarrow \mathbb{R}$ where $\mathcal{{U}} = \mathcal{U}^1 \times \mathcal{U}^2
\times, \ldots, \times \mathcal{U}^k$. The cost $\mathcal{J}^i$ acts on joint state $x_t$ and control ${u}_t \in {U}$ at each time step measuring the distances between agents, deviation from the reference path, and the change in the control input. Each agent $i$ follows a stochastic decentralized policy $\mathcal{F}^i_{\theta^i}:\mathcal{X} \longrightarrow \mathcal{U}^i$, which captures the probability of agent $i$ selecting action $u^i_t$ at time $t$, given that the system is in state $x_t$. 
The parameters for the policy $\mathcal{F}^i$ are represented by $\theta^i$, which may correspond to weights of a neural network or coefficients in a weighted linear function. We assume that we are provided with a dataset $\mathcal{\widehat D} = \{ \widehat \Gamma^{1:k}\}$ that consists of expert demonstration trajectories, $\widehat \Gamma^i$ for each agent $i$. Further, we let $\mathcal{D}_{\theta^i} = \{\Gamma^{1:k}_{\theta^i}\}$ denote the dataset of trajectories generated by the policy $\mathcal{F}^i_{\theta^i}$ parameterized by $\theta^i$.

\subsection{Background: MaxEnt IRL}

As background, we first briefly review the key details of (single-agent) MaxEnt IRL~\cite{singleirl, maxentirl}. As there is only one agent ($k=1$), we omit the superscript $i$ for our discussion on single-agent MaxEnt IRL. Given expert trajectories and assuming that the human is boundedly rational, the probability of a trajectory \( \Gamma \) under the MaxEnt IRL framework is defined as:

\[  P\left(\Gamma\right) \propto \exp\left(\sum_{t=0}^{T} \mathcal{J}\left(x_t, u_t\right)\right) \]

where \( \mathcal{J}(x_t, u_t) \) represents the cost at time \( t \) in joint state \( x_t \) and taking action \( u_t \). 

Let $\Phi(\cdot)$ be a feature function that, given a trajectory, returns a vector of features $\left[ \phi_1, \phi_2, \ldots, \phi_m \right]^\top$, averaged across the length of the trajectory, where $m$ refers to the number of distinct features. A simple example of a feature function is the weighted linear sum  $\Phi\left( \Gamma_{\theta}\right) = \sum_{t=0}^{T-1} {\theta}^\top \mathcal{J}\left( x_t, {u}_t \right)$. The objective in MaxEnt IRL is to adjust the weights of the cost function such that the expected features under the derived policy match the features of the expert, while also maximizing the entropy of the policy. The cost function is iteratively adjusted, typically using gradient-based methods, to ensure that the expected feature counts from the policy induced by the current reward function align with those from the expert demonstrations. To ease notation, let $\Psi\left( \Gamma_{\theta}, \widehat\Gamma \right) = \Phi\left( \Gamma_{\theta}\right) - \Phi\left(\widehat\Gamma\right)$ be a generalized discrepancy function. More formally, MaxEnt IRL finds $\theta^{*}$ that minimizes

\begin{equation}
        \left \lVert  \mathbb{E}_{\widehat\Gamma, \Gamma_{\theta}\sim\mathcal{\widehat D},\mathcal{D}_{\theta}} \Psi\left( \Gamma_{\theta}, \widehat\Gamma \right) \right \rVert
        \label{eq: maxentirl}
\end{equation}

There are several reasons why the MaxEnt IRL has been successful. First, the maximum entropy principle ensures a unique solution to the reward recovery problem, mitigating ambiguities inherent in traditional IRL. Further, resultant policies are inherently stochastic, capturing a broader range of behaviors and offering robustness in dynamic environments. Finally, MaxEnt IRL can generalize better to unseen states, making it particularly suitable for complex environments like pedestrian crowds~\cite{singleirl}.

\subsection{Problem Formulation}
We can generalize \eqref{eq: maxentirl} to the multi-agent setting.
Our problem statement is then,

\begin{problem}
    Find $\theta^{i,*}$ for $i \in [1,k]$ that minimizes
      \begin{equation}
        \left \lVert  \mathbb{E}_{\widehat\Gamma^i, \Gamma^i_{\theta^i}\sim\mathcal{\widehat D},\mathcal{D}_{\theta^i}} \Psi^i\left( \Gamma^i_{\theta^i}, \widehat\Gamma^i \right) \right \rVert
        \label{eq: unified_problem_statement}
    \end{equation}
\end{problem}




\noindent Note that we slightly abuse notation and ignore the terminal cost at $t=T$.

\section{Multi-agent MaxEnt IRL for Unstructued Pedestrian Crowds}
\label{sec: mairl_pedestrian_crowds}

We visually summarize the algorithm in Figure~\ref{fig: overview}. This section begins with a brief motivation for Multi-agent MaxEnt IRL, followed by describing the algorithm in detail, which includes the Tractability-Rationality trade-off trick which is depicted in the yellow box in Figure~\ref{fig: overview}.

In MaxEnt IRL~\cite{singleirl}, the goal is to learn a cost function that applies either to a single agent or is designed to collectively model a group of individuals. However, individuals in groups behave differently according to their nature. For instance, some individuals might be more risk averse than others, and therefore, are more likely to have a higher reward for collision avoidance. Others might place greater importance on reaching their goals faster and place greater weight for the cost of distance to goal. In~\cite{mairl}, Mehr et al. introduced the following multi-agent formulation of MaxEnt IRL for unstructured pedestrian crowds. 

Given a feature function  $\Phi^i\left( \Gamma^i_{\theta^i}\right)$,
that computes the features for agent $i$ with respect to weights $\theta^i$ and costs $\mathcal{J}^i\left( x_t, \mathbf{u}_t \right)$, solving Equation~\eqref{eq: unified_problem_statement} reduces to finding $\theta^{i,*} $ for every $i$ such that,

\begin{equation}
    \theta^{i,*} = \arg\min_{\theta^i}\left \lVert \mathbb{E}_{\Gamma^i_{\theta^i}\sim\mathcal{D}_{\theta^i}}\Phi^i\left( \Gamma^i_{\theta^i}\right) - \mathbb{E}_{\widehat\Gamma^i\sim\mathcal{\widehat D}}\Phi^i\left(\widehat\Gamma^i\right) \right \rVert
    \label{eq: problem_statement_sec4}
\end{equation}

Equation~\eqref{eq: problem_statement_sec4} can be efficiently solved via block coordinate descent~\cite{wright2015coordinate} where we treat $\theta$ as a concatenation of each $\theta^i$, where $i$ then refers to a coordinate. An update to the cost parameters at the current iteration, $\theta^i$, of agent $i$ is of the following form,

\begin{equation}
        \theta^i_\textrm{new} \gets \theta^i - \beta \left( \mathbb{E}_{\widehat\Gamma^i, \Gamma^i_{\theta^i}\sim\mathcal{\widehat D},\mathcal{D}_{\theta^i}}\Phi^i\left( \Gamma^i_{\theta^i}\right) - \Phi^i\left(\widehat\Gamma^i\right) \right)
    \label{eq: new_theta}
\end{equation}

where we compute the features for trajectories generated from the new set of cost parameters $\left(\theta^i_\textrm{new}\right)$ and proceed to update the cost parameters for the next agent. We iterate over each agent $i$ and repeat this process until convergence. At each iteration of the update process, each agent $i$ uses a dynamic game solver~\cite{le2022algames} as its parameterized stochastic policy $\mathcal{F}^i_{\theta^i}\left(x_t \right)$ to roll out $M$ trajectories $\{\Gamma^i\}_{1:M}$ corresponding to the updated cost parameters $\left(\theta^i_\textrm{new}\right)$. In the case of multi-agent IRL, the policy $\mathcal{F}^i_{\theta^i}$ must be obtained at each iteration using the current parameters.

Recently, Mehr et al.~\cite{mairl} showed that when the cost function for an agent, $\mathcal{J}^i$, assumes a quadratic form with linear system dynamics, then the set of policies, $\mathcal{F}^i_{\theta^i}\left(x_t \right)$, for $k$ boundedly rational agents can be obtained via backwards recursion through a set of Ricatti equations~\cite{lecleach2021lucidgames}. In unstructured crowds, however, the pedestrian dynamics are unknown. Recent IRL-based studies found that constant velocity~\cite{scholler2020constant} and constant acceleration~\cite{singleirl} motion models achieve a good enough approximation for pedestrian trajectory prediction benchmarks. Inspired by this result, we approximate the unknown systems dynamics with constant velocity dynamics,

\begin{align}
\left[\begin{array}{c}
\dot{p}^{i,x} \\
\dot{p}^{i,y} \\
\dot{\psi}^i
\end{array}\right]=\left[\begin{array}{ccc}
1& 0 & 0 \\
0& 1 & 0 \\
0& 0 & 0 \\
\end{array}\right]\left[\begin{array}{c}
v^{i,x} \\
v^{i,y}\\
0\\
\end{array}\right],
\label{eqn: constant_vel}
\end{align}

where $x^i_t = [p^{i,x}, p^{i,y}, \psi^i]^\top$ represents the 2D position and heading angle of agent $i$. Furthermore, as we do not know the cost functions $\mathcal{J}^i$, we generate a quadratic approximation via Taylor series expansion around the cumulative joint state value $\bar{x}_t = x_{1:t}$ to approximate $\mathcal{J}^i$ as

\begin{equation}
    \mathcal{\widetilde J}^i\left(\bar{x} + \delta_{x_t}\right) \approx \mathcal{\widetilde J}^i\left(\bar{x}\right) + \frac{1}{2} \delta_{x_t}^\top H^i_t \delta_{x_t} + {l^i_t}^\top \delta_{x_t}
    \label{eq: cost_approx}
\end{equation}

where $\delta_{x_t} = \bar{x}_t - x_t$ is the difference between the cumulative joint state and the joint state at the current time step. Note that this formulation
only considers nonlinear costs on state variables and the dependence on agents’ actions is quadratic. A similar approximation may be derived for a more general
case where the cost function is nonlinear in control as well. The matrix $H^i_t$ and vector $l^i_t$ correspond to the Hessian and gradient of the approximate cost function with respect to $x_t$. Each policy, $\mathcal{F}^i_{\theta^i}\left(x_t \right)$, is a Gaussian distribution where the mean and covariance parameters are functions of $H^i_t$:

\begin{equation}
\begin{split}
    \mathcal{F}^i_{\theta^i}\left(x_t \right) &= \mathcal{N}\left( \mu^i_t, \Sigma^i_t \right) \\
    \mu^i_t, \Sigma^i_t &\gets f\left ( H^i_t\right)^{-1} \\
\end{split}
\label{eq: H}
\end{equation}

where $f(\cdot)$ is a linear matrix transformation function (c.f.~\cite{mairl} for more details of $f(\cdot)$).

\paragraph*{\textbf{Intractability of Equation~\eqref{eq: cost_approx}}} 
Our assumption of constant velocity dynamics in Equation~\eqref{eqn: constant_vel}, combined with the fact that humans typically walk along linear paths at near identical speeds on average, results in a linear approximation of the cost function in Equation~\eqref{eq: cost_approx}. Under constant velocity dynamics and linear motion, the state evolution is linear. The deviations $\delta_{x_t}$ from the trajectory \(\bar{x}_t\) remain small and linear. Consequently, the second-order derivatives i.e., the entries of $\left (H_t^i\right)$, become very small since there is minimal curvature in the cost function along linear trajectories. Mathematically, the curvature of the cost function is captured by the eigenvalues of the Hessian $H_t^i$. If the cost function has little curvature (as it does under linear motion), these eigenvalues are close to zero, resulting in a non positive semi-definite (PSD) matrix.

\paragraph*{\textbf{Tractability-Rationality Trade-off Trick}} 

To achieve computational tractability, we relax the assumption of bounded rationality, which is typically represented by the stochastic nature of the agent's policy \(\mathcal{F}^i_{\theta^i}(x_t)\), characterized by the covariance matrix \(\Sigma^i_t\). The bounded rationality assumption implies that agents exhibit some randomness in their decision-making, often modeled as a probabilistic distribution over actions.

However, performing computations directly with \(\Sigma^i_t\) can lead to intractable solutions due to its dependence on $H_t^i$ which may be non-PSD. To address this, we modify the covariance matrix by adding a function of its diagonal elements:

\begin{equation}
\widetilde{\Sigma}^i_t = \Sigma^i_t + g\left(\textrm{diag}(\Sigma^i_t)\right) I,
\label{eq: trick}
\end{equation}
\begin{table*}[t]
\centering
\resizebox{\textwidth}{!}{
\begin{tabular}{rccccccc}
\toprule
Method & ETH & Hotel & Univ & Zara1 & Zara2 & Speedway & Average \\
\midrule
Social-GAN~\cite{gupta2018social} & $0.81/1.52$ & $0.72/1.61$ & $0.60/1.26$ & $0.34/0.69$ & $0.42/0.84$ &-& $0.58/1.18$ \\
SoPhie~\cite{sadeghian2019sophie} & $0.70/1.43$ & $0.76/1.67$ & $0.54/1.24$ & $0.30/0.63$ & $0.38/0.78$ &-& $0.54/1.15$ \\
CGNS~\cite{li2019cgns} & $0.62/1.40$ & $0.70/0.93$ & $0.48/1.22$ & $0.32/0.59$ & $0.35/0.71$ &-& $0.49/0.97$ \\
S-BiGAT*~\cite{Kosaraju2019BIGAT} & $0.69/1.29$ & $0.49/1.01$ & $0.55/1.32$ & $0.30/0.62$ & $0.36/0.75$ &-& $0.48/1.00$ \\
MATF~\cite{zhao2019multi} & $1.01/1.75$ & $0.43/0.80$ & $0.44/0.91$ & $0.26/0.45$ & $0.26/0.57$ &-& $0.48/0.90$ \\
Next*~\cite{liang2019peeking} & $0.73/1.65$ & $0.30/0.59$ & $0.60/1.27$ & $0.38/0.81$ & $0.31/0.68$ &-& $0.46/1.00$ \\
SR-LSTM~\cite{zhang2019sr} & $0.63/1.25$ & $0.37/0.74$ & $0.51/1.10$ & $0.41/0.90$ & $0.32/0.70$ &-& $0.45/0.94$ \\
STGAT~\cite{Huang2019stgat} & $0.65/1.12$ & $0.35/0.66$ & $0.52/1.10$ & $0.34/0.69$ & $0.29/0.60$ &-& $0.43/0.83$ \\
GOAL-GAN*~\cite{Dendorfer_2020_ACCV} & $0.59/1.18$ & $0.19/0.35$ & $0.60/1.19$ & $0.43/0.87$ & $0.32/0.65$ &-& $0.43/0.85$ \\
SGCN~\cite{shi2021sgcn} & $0.63/1.03$ & $0.32/0.55$ & $0.37/0.70$ & $0.29/0.53$ & $0.25/0.45$ &-& $0.37/0.65$ \\
MANTRA~\cite{marchetti2020memnet} & $0.48/0.88$ & $0.17/0.33$ & $0.37/0.81$ & $0.27/0.58$ & $0.30/0.67$ &-& $0.32/0.65$ \\
Transformer~\cite{giuliari2020transformer} & $0.61/1.12$ & $0.18/0.30$ & $0.35/0.65$ & $0.22/0.38$ & $0.17/0.32$ & -&$0.31/0.55$ \\
SMEMO~\cite{marchetti2024smemo} & $0.39/0.59$ & $0.14/0.20$ & $0.23/0.41$ & $0.19/0.32$ & $0.15/0.26$ &-& $0.22/0.35$ \\
Introvert~\cite{shafiee2021introvert} & $0.42/0.70$ & $0.11/0.17$ & $0.20/0.32$ & $0.16/0.27$ & $0.16/0.25$ &-& $0.21/0.34$ \\
PCCSNet~\cite{sun2021three} & $0.28/0.54$ & $0.11/0.19$ & $0.29/0.60$ & $0.21/0.44$ & $0.15/0.34$ &-& $0.21/0.42$ \\
\midrule
MaxEnt IRL~\cite{singleirl}         & $0.76$ / $1.48$ & $0.68$ / $1.74$ & $0.51$ / $1.21$ & $0.37$ / $0.69$ & $0.29$ / $0.75$ &$0.40/0.80$& $0.50/1.11$ \\
PECNet~\cite{mangalam2020not} & $0.54/0.87$ & $0.18/0.24$ & $0.35/0.60$ & $0.22/0.39$ & $0.17/0.30$ &$0.35/0.75$& $0.30/0.51$ \\
SceneTransformer~\cite{ngiam2021scene} &  $0.50$ / $0.76$ & $0.14$ / $0.20$ & $0.29$ / $0.42$ & $0.22$ / $0.36$ & $0.16$ / $0.27$ & $0.33/0.72$ & $0.27/0.46$\\
AgentFormer~\cite{yuan2021agentformer} & $0.45/0.75$ & $0.14/0.22$ & $0.25/0.45$ & $0.18/0.30$ & $0.14/0.24$ &$0.31/0.76$ & $0.24/0.44$ \\
HST~\cite{salzmann2023robots} & $0.41$ / $0.73$ & \textbf{$0.10$ / $0.14$} & $0.24$ / $0.44$ & $0.17$ / $0.30$ & $0.14$ / $0.24$ & $0.35/0.68$ & $0.24/0.42$\\
LB-EBM~\cite{pang2021trajectory} & $0.30/0.52$ & $0.13/0.20$ & $0.27/0.52$ & $0.20/0.37$ & $0.15/0.29$ & $0.32/0.81$& $0.23/0.45$ \\
Trajectron++~\cite{salzmann2020trajectron++} & $0.39/0.83$ & $0.12/0.19$ & $0.22/0.43$ & $0.17/0.32$ & $0.12/0.25$ &$0.29/0.75$ & $0.22/0.47$ \\
Expert-Goals~\cite{he2021where} & $0.37/0.65$ & $0.11/0.15$ & $0.20/0.44$ & $0.15/0.31$ & $0.12/0.25$ &$0.29/0.69$& $\mathbf{0.21/0.41}$ \\
\midrule
MAIRL & {$0.39$ / $0.71$} & $0.13$ / $0.18$ & {$0.23$ / $0.32$} & {$0.19$ / $0.29$} & {$0.15$ / $0.28$} & $\mathbf{0.26/0.60}$ & $0.23$ / $0.39$ \\

\bottomrule
\end{tabular}
}
\caption{Quantitative results (ADE/FDE) of our proposed method compared to state-of-the-art baselines on five benchmark datasets. All results are in meters. Lower is better}
\label{tab: big_table}
\end{table*}

where \(g\left(\textrm{diag}(\Sigma^i_t)\right)\) produces a vector that adjusts the diagonal elements of \(\Sigma^i_t\). This operation ensures that \(\widetilde{\Sigma}^i_t\) is a positive semi-definite (PSD) matrix, thus preserving the stochastic nature of the policy while making the problem more tractable. The diagonal elements of \(\Sigma^i_t\), denoted as \(\sigma^{ij}, i=j\), quantify the uncertainty or randomness in the action that agent \(i\) takes toward agent \(j\). By applying the function \(g\) to these diagonal elements, we essentially decrease the variance (uncertainty) of the agent's actions. This modification shifts towards a more tractable model at the cost of potentially capturing a less accurate representation of agent behaviors, as seen in our evaluation.

\noindent\textit{Remark:} There are several new open problems in this research direction. First, in this work, we heuristically select $g(\cdot)$ from multiple viable candidates $\left (g\left(\textrm{diag}\left(\Sigma^i_t\right)\right) I = I\right)$ is a trivial choice which enforces tractability but eliminates rationality). Establishing a suitable class of functions for $g(\cdot)$ remains a topic for future work. Another interesting open problem is to establish a bound on the change in the agents' behavior due to the adjustment of the covariance matrix, \(\widetilde{\Sigma}^i_t\), and its subsequent impact on the mixed-Nash equilibrium.

\begin{table*}[t]
    \centering
    \resizebox{\textwidth}{!}{
    \begin{tabular}{lcccccccc}
        \toprule
        \textbf{Dataset} & PecNET~\cite{pecnet} & SMEMO~\cite{marchetti2024smemo} & OpenTraj~\cite{amirian2020opentraj} & JRDB\_Traj~\cite{saadatnejad2023jrdb} & MaxEnt IRL~\cite{singleirl}& Social-pose~\cite{gao2022social} & HST~\cite{salzmann2023robots} & MAIRL \\
        \midrule
    JRDB~\cite{jrdb}        &$3.939	$ & $2.671$ & $3.498$& $2.646$ & $4.249$ & $2.558$& $\mathbf{2.506}$ & $2.640$ \\
    SCAND~\cite{scand}                     & $0.55 / 1.20$ &$ 0.50 / 1.10$ & $0.44 / 0.90$ & $0.36 / 0.81$ & $0.42 / 0.85$ & - &-& $\mathbf{0.23/0.67}$\\
        \bottomrule
    \end{tabular}
    }
    \caption{EFE (End-to-end Forecasting Error) results on the JRDB and SCAND dataset. All results are in meters/ Lower is better.}
    \label{tab: small_table}
\end{table*}




\section{Evaluation}
\label{sec: experiments}

\begin{figure}[t]
\centering

   \begin{subfigure}[h]{.49\columnwidth}
    \includegraphics[width=\textwidth]{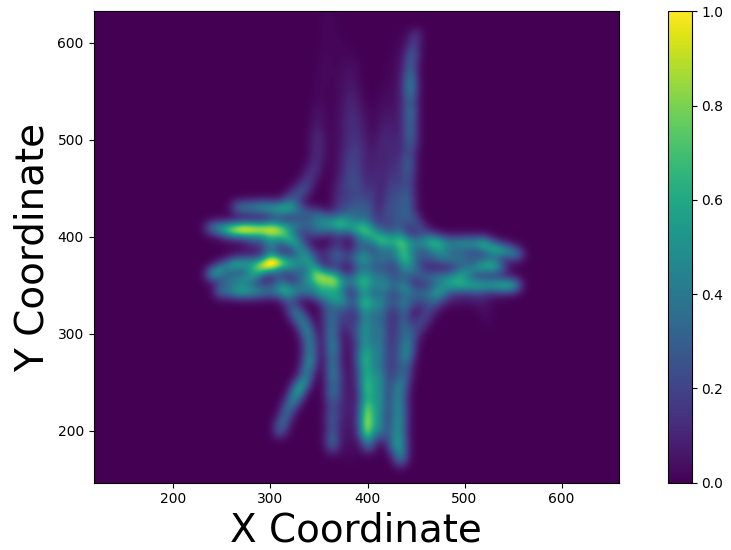}
    \caption{Our dataset}
    \label{fig: lines_ped}
  \end{subfigure}
   \begin{subfigure}[h]{.49\columnwidth}
    \includegraphics[width=\textwidth]{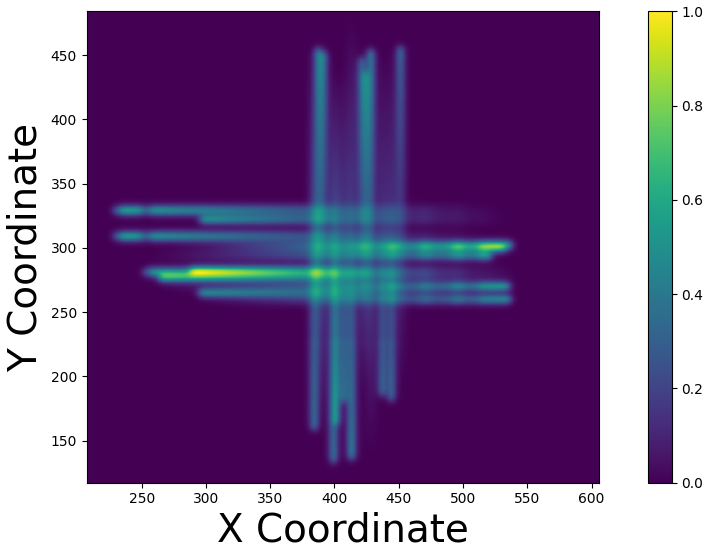}
    \caption{Waymo/INTERACTION}
    \label{fig: lines_veh}
  \end{subfigure}
\caption{Visualizing and comparing the entropy of trajectories in the Speedway dataset ($1.061$ bits) with the entropy of trajectories in the Waymo and INTERACTION datasets ($0.336$ and $0.475$ bits, respectively). We observe that trajectories in the Speedway dataset are denser, more unstructured, and have a higher entropy.} 
  \label{fig: lines}
  \vspace{-10pt}
\end{figure} 
We compare our multi-agent MaxEnt IRL approach with single-agent MaxEnt IRL and state-of-the-art trajectory planning and prediction approaches and aim to understand how our method compares to these baselines on popular autonomous driving and pedestrian crowd datasets.

\subsection{Data Pre-processing}



We collected an unstructured real world pedestrian crowd dataset at one of the busiest intersections on a University campus using a BlueCity\footnote{\href{https://ouster.com/products/software/blue-city}{https://ouster.com/products/software/blue-city}} Velodyne LiDAR mounted at the intersection. Our dataset consists of more tightly coupled agent interactions compared to existing motion planning datasets~\cite{scand, jrdb, waymo}. 

The BlueCity LiDAR is equipped with in-built object detection and tracking modules~\cite{yolo}. The LiDAR continuously streams data as frames where each frame consists of the timestamp and a sequence of frame objects corresponding to agents observed by the LiDAR. Each frame object consists of a unique id that is persistent in all the frames in which the agent appears, location of the agent in meter with respect to its distance to the sensor, width and length of the bounding box detected for the agent, the angle of the detected bounding box in radian, the class type of the agent such as `pedestrian' or `car', the speed of the agents, and the accuracy of the detection. We recorded trajectories consisting of the id of the agents, their locations, speed, and orientation during busy moments (such as in between classes). 

We processed the data by removing agents that were either at a standstill or appeared at the edges of the LiDAR's field of view. Furthermore, we clipped each trajectory within a uniform spatio-temporal range ($[-20m, +20m]$ in the x axis and $[-10m, 15m]$ in the y axis). After processing the raw trajectories, we ended up with $20$ trajectories of pedestrians with $5$ trajectories each going from east to west (denoted as `W'), west to east (denoted as `E'), north to south (denoted as `S'), south to north (denoted as `N'). We created $4$ different categories, `W-E-S',`W-E-N',`S-N-W',`S-N-E', with $5^3 = 125$ trajectories each for a total of $500$ trajectories.

Each trajectory is finally processed in the form of a $T \times 4k$ array where $T$ corresponds to the number of frames in that interaction and $4k$ corresponds to the dimension of the joint state, that is, $x_t = \left[ p^i, v^i, \phi^i \right]^\top$ for $i \in [1,k], p^i \in \mathbb{R}^2, v^i\in \mathbb{R}^1, \phi^i \in \mathbb{S}^1$. The final processed trajectories in the speedway dataset form the demonstration trajectories, denoted as $\mathcal{\widehat D}$. Features are computed based on distance to the goal position, distance to other agents, and control effort.

\subsection{Baselines and Evaluation Metric}
We compare our approach, which we call MAIRL (multi-agent IRL), with state-of-the-art trajectory prediction methods that have consistently performed well on leading benchmarks including the JRDB and ETH/UCY.
To compare MAIRL with these methods\footnote{Note that for comparison on the Speedway dataset, baselines were selected by identifying state-of-the-art methods that $(i)$ were compatible with our input data format and $(ii)$ had open source implementations\label{fnlabel}.} on the Speedway dataset, we directly use the opensource software for these baselines and train them on our dataset to present a direct evaluation. We set the learning rate to $3 \times 10^{-4}$ and perform mini-batch training with a batch size of $64$. A $60-40$ cross-validation split is used for evaluation.  To prevent overfitting, we apply a dropout rate of $0.2$. Training utilizes the AdamW optimizer with the Mean Squared Error (MSE) loss function. All training was performed on an NVIDIA GeForce RTX 2080 GPU machine and the average time to train the baselines was in the order of minutes. 

Following standard evaluation procedures followed by the literature~\cite{singleirl, mairl}, we measure the average displacement error (ADE) and the final displacement error (FDE) between the demonstration trajectories and the trajectories generated via the policy $\mathcal{F}^i_{\theta^i}$. 

\begin{figure}[t]
\centering
   \begin{subfigure}[h]{0.235\textwidth}
    \includegraphics[width=\textwidth]{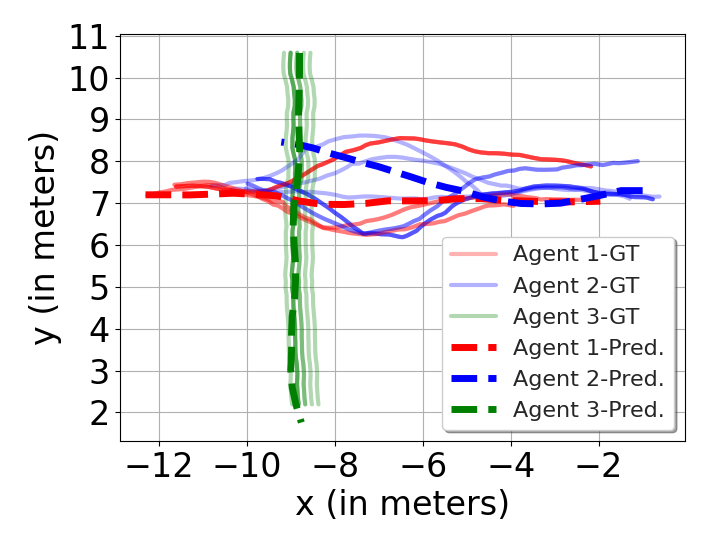}
    \caption{Multi-agent Max-Ent IRL}
    \label{fig: mairl}
  \end{subfigure}
   \begin{subfigure}[h]{0.235\textwidth}
    \includegraphics[width=\textwidth]{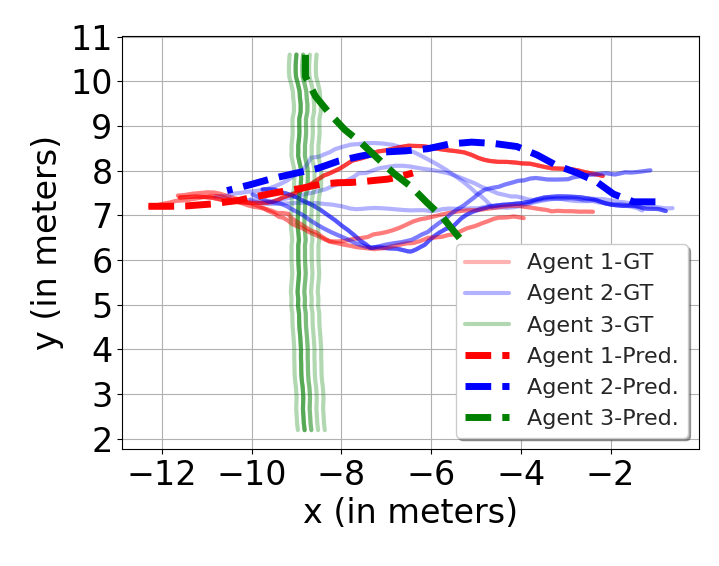}
    \caption{Max-Ent IRL~\cite{maxentirl}}
    \label{fig: irl}
  \end{subfigure}
\caption{\textbf{Qualitative comparison}--Each color represents a pedestrian. Solid faded lines represent demonstration trajectories and dashed lines represent trajectories generated from the learned policies. The directions of movement for the pedestrians are north to south (\textcolor{ForestGreen}{$\mathbf{\downarrow}$}), east to west ($\mathbf{\textcolor{blue}{\leftarrow}}$) and west to east ($\mathbf{\textcolor{red}{\rightarrow}}$). `GT' and  `Pred' refer to ground truth and predicted trajectories. We inspect how closely the predicted trajectories align with the ground truth distribution.  
} 
  \label{fig: main_results}
\end{figure}  

\begin{figure}[t]
\centering
  %
    \includegraphics[width=0.48\textwidth]{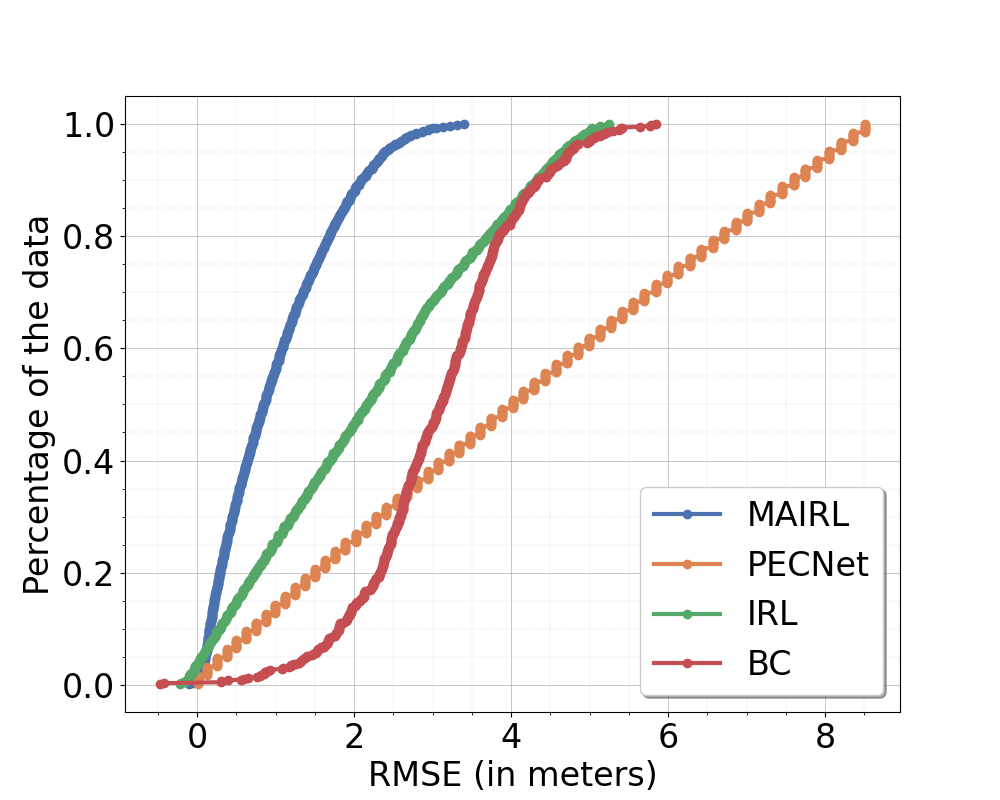}
    \caption{Cumulative RMSE distribution demonstrates the percentage of trajectories below a certain RMSE threshold; steeper curves indicate a more effective learner.}
    \label{fig: cdf}

\end{figure}  

\subsection{Results and Discussion}


Table~\ref{tab: big_table} compares (in order of better performance) the ADE/FDE (lower is better) of MAIRL with state-of-the-art trajectory prediction and planning methods on the well-known ETH/UCY datasets as well as on Speedway, in meters. The horizontal rule in the middle of the table divides the methods between ones that could be tested on the Speedway dataset versus ones that could not\footref{fnlabel}. The results show that our MAIRL outperforms the top methods on the Speedway dataset, and is comparable on the ETH/UCY datasets, only performing worse than Expert-Goals~\cite{he2021where} and Trajectron++~\cite{salzmann2020trajectron++}, two well-adopted methods for deep learning-based trajectory prediction. Table~\ref{tab: small_table} additionally compares MAIRL on the JRDB and SCAND datasets based on the End-to-End Forecasting Error (EFE), a specific metric used by the JRDB benchmakr, but is similar to the ADE/FDE metrics, again, in meters. MAIRL performs the best on SCAND, and comes in $3^{\textrm{rd}}$ on JRDB, behind Social-Pose~\cite{gao2022social} and HST~\cite{salzmann2023robots}, two methods that additionally take into account the skeleton pose of the pedestrians. In both Tables~\ref{tab: big_table} and~\ref{tab: small_table}, note that we also compare MAIRL with single agent MaxEnt IRL, which performs considerably worse.

Figure~\ref{fig: main_results} visually illustrates the comparison between multi- and singe-agent MaxEnt IRL, where solid faded lines represent the ground truth demonstration trajectories and dashed lines represent the predicted trajectories. The visualization corresponds to the intersection data comprising of $3$ agents, each representing a different color. The directions of movement for the pedestrians are north to south (\textcolor{ForestGreen}{$\mathbf{\downarrow}$}), east to west ($\mathbf{\textcolor{blue}{\leftarrow}}$) and west to east ($\mathbf{\textcolor{red}{\rightarrow}}$). We inspect how closely the predicted trajectories are aligned with the ground truth distribution. Due to the single-agent nature of MaxEnt IRL, it aligns the predicted trajectories with the ground truth demonstrations for at most $1$ agent (the blue agent), and fails for the green (extreme deviation) and the red (not reached goal) agents. Note that even though MAIRL outperforms all three baselines in terms of aligning with the ground truth demonstrations, it still remains imperfect in its own alignment due to the reduction in accuracy resulting from the Tractability-Rationality Trade-off, as discussed in Section~\ref{sec: problem_formulation}.

In Figure~\ref{fig: cdf}, we plot a cumulative distribution frequency (CDF) that measures the percentage of trajectories that fall below a certain error threshold; steeper curves imply a more effective algorithm as it indicates that that the algorithm was able to generate more low-ADE trajectories. The CDF plot is useful because the ADE/FDE metrics alone may not convey the full picture. For example, a baseline could yield a poor ADE, and be declared as weak algorithm, but a low ADE could result from one or two outlier trajectories with an particularly high ADE, thereby skewing the average against its favor.

\section{Conclusion and Open Questions}
\label{sec: conclusion}

We show that multi-agent MaxEnt inverse reinforcement learning provides accurate predictions of pedestrians’ motion in real world unstructured pedestrian crowds with only few training trajectories. Our key insight was a mathematical trick that enabled tractability of multi-agent MaxEnt IRL at the cost of a slight reduction in accuracy. We used a dynamic game theoretic solver to compute optimal policies that maximize the reward functions learned via our approach. We evaluated this new approach using a dataset collected at a busy intersection on a University campus and showed that despite the accuracy trade-off, our approach outperforms state-of-the-art imitation learning baselines such as single-agent MaxEnt IRL, behavior cloning, and generative modeling. Our findings highlighted that our approach yields the lowest error between the policy-generated trajectories. Future work includes finding ways to solve Equation~\eqref{eq: H} in unstructured pedestrian crowds without relaxing bounded rationality. Another open problem is to compute an exact bound for the tractability-rationality tradeoff trick (Section~\ref{sec: mairl_pedestrian_crowds}). In addition, incorporating external factors like social contexts and the environment in modeling pedestrian behaviors can yield more holistic and precise predictions, providing a more comprehensive view of motion dynamics in various environments.
\bibliography{refs}

\begin{thebibliography}{10}

\bibitem{chandra2024deadlock}
R.~Chandra, V.~Zinage, E.~Bakolas, P.~Stone, and J.~Biswas, ``Deadlock-free, safe, and decentralized multi-robot navigation in social mini-games via discrete-time control barrier functions,'' 2024.

\bibitem{chandra2022game}
R.~Chandra, M.~Wang, M.~Schwager, and D.~Manocha, ``Game-theoretic planning for autonomous driving among risk-aware human drivers,'' in {\em 2022 International Conference on Robotics and Automation (ICRA)}, pp.~2876--2883, 2022.

\bibitem{chandra2022gameplan}
R.~Chandra and D.~Manocha, ``Gameplan: Game-theoretic multi-agent planning with human drivers at intersections, roundabouts, and merging,'' {\em IEEE Robotics and Automation Letters}, vol.~7, no.~2, pp.~2676--2683, 2022.

\bibitem{suriyarachchi2022gameopt}
N.~Suriyarachchi, R.~Chandra, J.~S. Baras, and D.~Manocha, ``Gameopt: Optimal real-time multi-agent planning and control at dynamic intersections,'' in {\em 2022 IEEE 25th International Conference on Intelligent Transportation Systems (ITSC)}, pp.~2599--2606, IEEE doi - 10.1109/ITSC55140.2022.9921968, 2022.

\bibitem{mavrogiannis2022b}
A.~Mavrogiannis, R.~Chandra, and D.~Manocha, ``B-gap: Behavior-rich simulation and navigation for autonomous driving,'' {\em IEEE Robotics and Automation Letters}, vol.~7, no.~2, pp.~4718--4725, 2022.

\bibitem{scand}
H.~Karnan, A.~Nair, X.~Xiao, G.~Warnell, S.~Pirk, A.~Toshev, J.~Hart, J.~Biswas, and P.~Stone, ``Socially compliant navigation dataset (scand): A large-scale dataset of demonstrations for social navigation,'' {\em IEEE Robotics and Automation Letters}, vol.~7, no.~4, pp.~11807--11814, 2022.

\bibitem{raj2023targeted}
A.~H. Raj, Z.~Hu, H.~Karnan, R.~Chandra, A.~Payandeh, L.~Mao, P.~Stone, J.~Biswas, and X.~Xiao, ``Targeted learning: A hybrid approach to social robot navigation,'' 2023.

\bibitem{sprague2023socialgym}
Z.~Sprague, R.~Chandra, J.~Holtz, and J.~Biswas, ``Socialgym 2.0: Simulator for multi-agent social robot navigation in shared human spaces,'' {\em arXiv preprint arXiv:2303.05584}, 2023.

\bibitem{chandra2023socialmapf}
R.~Chandra, R.~Maligi, A.~Anantula, and J.~Biswas, ``Socialmapf: Optimal and efficient multi-agent path finding with strategic agents for social navigation,'' {\em IEEE Robotics and Automation Letters}, 2023.

\bibitem{chandra2023decentralized}
R.~Chandra, R.~Menon, Z.~Sprague, A.~Anantula, and J.~Biswas, ``Decentralized social navigation with non-cooperative robots via bi-level optimization,'' {\em arXiv preprint arXiv:2306.08815}, 2023.

\bibitem{smg}
R.~Chandra, V.~Zinage, E.~Bakolas, J.~Biswas, and P.~Stone, ``Decentralized multi-robot social navigation in constrained environments via game-theoretic control barrier functions,'' {\em arXiv preprint arXiv:2308.10966}, 2023.

\bibitem{poddar2023crowd}
S.~Poddar, C.~Mavrogiannis, and S.~S. Srinivasa, ``From crowd motion prediction to robot navigation in crowds,'' 2023.

\bibitem{francis2023principles}
A.~Francis, C.~Pérez-D'Arpino, C.~Li, F.~Xia, A.~Alahi, R.~Alami, A.~Bera, A.~Biswas, J.~Biswas, R.~Chandra, H.-T.~L. Chiang, M.~Everett, S.~Ha, J.~Hart, J.~P. How, H.~Karnan, T.-W.~E. Lee, L.~J. Manso, R.~Mirksy, S.~Pirk, P.~T. Singamaneni, P.~Stone, A.~V. Taylor, P.~Trautman, N.~Tsoi, M.~Vázquez, X.~Xiao, P.~Xu, N.~Yokoyama, A.~Toshev, and R.~Martín-Martín, ``Principles and guidelines for evaluating social robot navigation algorithms,'' 2023.

\bibitem{gupta2018social}
A.~Gupta, J.~Johnson, L.~Fei-Fei, S.~Savarese, and A.~Alahi, ``Social gan: Socially acceptable trajectories with generative adversarial networks,'' in {\em Proceedings of the IEEE Conference on Computer Vision and Pattern Recognition}, pp.~2255--2264, 2018.

\bibitem{sadeghian2019sophie}
A.~Sadeghian, V.~Kosaraju, A.~Sadeghian, N.~Hirose, H.~Rezatofighi, and S.~Savarese, ``Sophie: An attentive gan for predicting paths compliant to social and physical constraints,'' in {\em Proceedings of the IEEE Conference on Computer Vision and Pattern Recognition}, pp.~1349--1358, 2019.

\bibitem{Kosaraju2019BIGAT}
V.~Kosaraju, A.~Sadeghian, R.~Martin-Martin, I.~Reid, H.~Rezatofighi, and S.~Savarese, ``Social-bigat: Multimodal trajectory forecasting using bicycle-gan and graph attention networks,'' in {\em Advances in Neural Information Processing Systems}, vol.~32, Curran Associates, Inc., 2019.

\bibitem{giuliari2020transformer}
F.~Giuliari, I.~Hasan, M.~Cristani, and F.~Galasso, ``Transformer networks for trajectory forecasting,'' {\em arXiv preprint arXiv:2003.08111}, 2020.

\bibitem{marchetti2024smemo}
F.~Marchetti, F.~Becattini, L.~Seidenari, and A.~Del~Bimbo, ``Smemo: social memory for trajectory forecasting,'' {\em IEEE Transactions on Pattern Analysis and Machine Intelligence}, 2024.

\bibitem{sun2021three}
J.~Sun, Y.~Li, H.-S. Fang, and C.~Lu, ``Three steps to multimodal trajectory prediction: Modality clustering, classification and synthesis,'' {\em arXiv preprint arXiv:2103.07854}, 2021.

\bibitem{ngiam2021scene}
J.~Ngiam, B.~Caine, V.~Vasudevan, Z.~Zhang, H.-T.~L. Chiang, J.~Ling, R.~Roelofs, A.~Bewley, C.~Liu, A.~Venugopal, {\em et~al.}, ``Scene transformer: A unified architecture for predicting multiple agent trajectories,'' {\em arXiv preprint arXiv:2106.08417}, 2021.

\bibitem{yuan2021agentformer}
Y.~Yuan, X.~Weng, Y.~Ou, and K.~Kitani, ``Agentformer: Agent-aware transformers for socio-temporal multi-agent forecasting,'' {\em arXiv preprint arXiv:2103.14023}, 2021.

\bibitem{salzmann2023robots}
T.~Salzmann, H.-T.~L. Chiang, M.~Ryll, D.~Sadigh, C.~Parada, and A.~Bewley, ``Robots that can see: Leveraging human pose for trajectory prediction,'' {\em IEEE Robotics and Automation Letters}, 2023.

\bibitem{salzmann2020trajectron++}
T.~Salzmann, B.~Ivanovic, P.~Chakravarty, and M.~Pavone, ``Trajectron++: Multi-agent generative trajectory forecasting with heterogeneous data for control,'' {\em arXiv preprint arXiv:2001.03093}, 2020.

\bibitem{he2021where}
Z.~He and R.~P. Wildes, ``Where are you heading? dynamic trajectory prediction with expert goal examples,'' in {\em Proceedings of the International Conference on Computer Vision (ICCV)}, Oct. 2021.

\bibitem{waymo}
S.~Ettinger, S.~Cheng, B.~Caine, C.~Liu, H.~Zhao, S.~Pradhan, Y.~Chai, B.~Sapp, C.~R. Qi, Y.~Zhou, {\em et~al.}, ``Large scale interactive motion forecasting for autonomous driving: The waymo open motion dataset,'' in {\em Proceedings of the IEEE/CVF International Conference on Computer Vision}, pp.~9710--9719, 2021.

\bibitem{singleirl}
D.~Gonon and A.~Billard, ``Inverse reinforcement learning of pedestrian--robot coordination,'' {\em IEEE Robotics and Automation Letters}, 2023.

\bibitem{maxentirl}
B.~D. Ziebart, A.~L. Maas, J.~A. Bagnell, A.~K. Dey, {\em et~al.}, ``Maximum entropy inverse reinforcement learning.,'' in {\em Aaai}, vol.~8, pp.~1433--1438, Chicago, IL, USA, 2008.

\bibitem{natarajan2010multi}
S.~Natarajan, G.~Kunapuli, K.~Judah, P.~Tadepalli, K.~Kersting, and J.~Shavlik, ``Multi-agent inverse reinforcement learning,'' in {\em 2010 ninth international conference on machine learning and applications}, pp.~395--400, IEEE, 2010.

\bibitem{lin2019multi}
X.~Lin, S.~C. Adams, and P.~A. Beling, ``Multi-agent inverse reinforcement learning for certain general-sum stochastic games,'' {\em Journal of Artificial Intelligence Research}, vol.~66, pp.~473--502, 2019.

\bibitem{yu2019multi}
L.~Yu, J.~Song, and S.~Ermon, ``Multi-agent adversarial inverse reinforcement learning,'' in {\em International Conference on Machine Learning}, pp.~7194--7201, PMLR, 2019.

\bibitem{song2018multi}
J.~Song, H.~Ren, D.~Sadigh, and S.~Ermon, ``Multi-agent generative adversarial imitation learning,'' {\em Advances in neural information processing systems}, vol.~31, 2018.

\bibitem{schwarting2019social}
W.~Schwarting, A.~Pierson, J.~Alonso-Mora, S.~Karaman, and D.~Rus, ``Social behavior for autonomous vehicles,'' {\em Proc. Nat. Acad. Sci.}, vol.~116, no.~50, pp.~24972--24978, 2019.

\bibitem{lecleach2021lucidgames}
S.~Lecleac'h, M.~Schwager, and Z.~Manchester, ``Lucidgames: Online unscented inverse dynamic games for adaptive trajectory prediction and planning,'' {\em IEEE Robot. Automat. Lett.}, vol.~6, pp.~5485--5492, Jul. 2021.

\bibitem{rothfuss2017inverse}
S.~Rothfuß, J.~Inga, F.~Köpf, M.~Flad, and S.~Hohmann, ``Inverse optimal control for identification in non-cooperative differential games,'' in {\em IFAC-PapersOnLine}, vol.~50, pp.~14909--14915, 2017.

\bibitem{kopf2017inverse}
F.~Köpf, J.~Inga, S.~Rothfuß, M.~Flad, and S.~Hohmann, ``Inverse reinforcement learning for identification in linear-quadratic dynamic games,'' in {\em IFAC-PapersOnLine}, vol.~50, pp.~14902--14908, 2017.

\bibitem{mairl}
N.~Mehr, M.~Wang, M.~Bhatt, and M.~Schwager, ``Maximum-entropy multi-agent dynamic games: Forward and inverse solutions,'' {\em IEEE Transactions on Robotics}, 2023.

\bibitem{kalman}
R.~E. Kalman, ``When is a linear control system optimal?,'' 1964.

\bibitem{abbeel2004apprenticeship}
P.~Abbeel and A.~Y. Ng, ``Apprenticeship learning via inverse reinforcement learning,'' in {\em Proceedings of the twenty-first international conference on Machine learning}, p.~1, 2004.

\bibitem{ng2000algorithms}
A.~Y. Ng, S.~Russell, {\em et~al.}, ``Algorithms for inverse reinforcement learning.,'' in {\em Icml}, vol.~1, p.~2, 2000.

\bibitem{sociallstm}
A.~Alahi, K.~Goel, V.~Ramanathan, A.~Robicquet, L.~Fei-Fei, and S.~Savarese, ``Social lstm: Human trajectory prediction in crowded spaces,'' in {\em Proceedings of the IEEE conference on computer vision and pattern recognition}, pp.~961--971, 2016.

\bibitem{chandra2019robusttp}
R.~Chandra, U.~Bhattacharya, C.~Roncal, A.~Bera, and D.~Manocha, ``Robusttp: End-to-end trajectory prediction for heterogeneous road-agents in dense traffic with noisy sensor inputs,'' in {\em Proceedings of the 3rd ACM Computer Science in Cars Symposium}, pp.~1--9, 2019.

\bibitem{chandra2019traphic}
R.~Chandra, U.~Bhattacharya, A.~Bera, and D.~Manocha, ``Traphic: Trajectory prediction in dense and heterogeneous traffic using weighted interactions,'' in {\em Proceedings of the IEEE/CVF Conference on Computer Vision and Pattern Recognition}, pp.~8483--8492, 2019.

\bibitem{Zhao2020TNTTT}
H.~Zhao, J.~Gao, T.~Lan, C.~Sun, B.~Sapp, B.~Varadarajan, Y.~Shen, Y.~Shen, Y.~Chai, C.~Schmid, C.~Li, and D.~Anguelov, ``Tnt: Target-driven trajectory prediction,'' in {\em Conference on Robot Learning}, 2020.

\bibitem{Lee2017DESIREDF}
N.~Lee, W.~Choi, P.~Vernaza, C.~B. Choy, P.~H.~S. Torr, and M.~Chandraker, ``Desire: Distant future prediction in dynamic scenes with interacting agents,'' {\em 2017 IEEE Conference on Computer Vision and Pattern Recognition (CVPR)}, pp.~2165--2174, 2017.

\bibitem{precog}
N.~Rhinehart, R.~Mcallister, K.~Kitani, and S.~Levine, ``Precog: Prediction conditioned on goals in visual multi-agent settings,'' in {\em 2019 IEEE/CVF International Conference on Computer Vision (ICCV)}, (Los Alamitos, CA, USA), pp.~2821--2830, IEEE Computer Society, nov 2019.

\bibitem{yoo2022gin}
S.-W. Yoo, C.~Kim, J.~Choi, S.-W. Kim, and S.-W. Seo, ``Gin: Graph-based interaction-aware constraint policy optimization for autonomous driving,'' {\em IEEE Robotics and Automation Letters}, vol.~8, no.~2, pp.~464--471, 2022.

\bibitem{cao2022leveraging}
Z.~Cao, E.~Biyik, G.~Rosman, and D.~Sadigh, ``Leveraging smooth attention prior for multi-agent trajectory prediction,'' in {\em 2022 International Conference on Robotics and Automation (ICRA)}, pp.~10723--10730, IEEE, 2022.

\bibitem{qi2018intent}
S.~Qi and S.-C. Zhu, ``Intent-aware multi-agent reinforcement learning,'' in {\em 2018 IEEE international conference on robotics and automation (ICRA)}, pp.~7533--7540, IEEE, 2018.

\bibitem{wang2021tom2c}
Y.~Wang, F.~Zhong, J.~Xu, and Y.~Wang, ``Tom2c: Target-oriented multi-agent communication and cooperation with theory of mind,'' {\em arXiv preprint arXiv:2111.09189}, 2021.

\bibitem{rabinowitz2018machine}
N.~Rabinowitz, F.~Perbet, F.~Song, C.~Zhang, S.~A. Eslami, and M.~Botvinick, ``Machine theory of mind,'' in {\em International conference on machine learning}, pp.~4218--4227, PMLR, 2018.

\bibitem{vemula2018social}
A.~Vemula, K.~Muelling, and J.~Oh, ``Social attention: Modeling attention in human crowds,'' in {\em 2018 IEEE international Conference on Robotics and Automation (ICRA)}, pp.~4601--4607, IEEE, 2018.

\bibitem{dai2023socially}
Z.~Dai, T.~Zhou, K.~Shao, D.~H. Mguni, B.~Wang, and H.~Jianye, ``Socially-attentive policy optimization in multi-agent self-driving system,'' in {\em Conference on Robot Learning}, pp.~946--955, PMLR, 2023.

\bibitem{wu2023multiagent}
H.~Wu, P.~Sequeira, and D.~V. Pynadath, ``Multiagent inverse reinforcement learning via theory of mind reasoning,'' {\em arXiv preprint arXiv:2302.10238}, 2023.

\bibitem{he2016opponent}
H.~He, J.~Boyd-Graber, K.~Kwok, and H.~Daum{\'e}~III, ``Opponent modeling in deep reinforcement learning,'' in {\em International conference on machine learning}, pp.~1804--1813, PMLR, 2016.

\bibitem{zhu2020multi}
Z.~Zhu, E.~B{\i}y{\i}k, and D.~Sadigh, ``Multi-agent safe planning with gaussian processes,'' in {\em 2020 IEEE/RSJ International Conference on Intelligent Robots and Systems (IROS)}, pp.~6260--6267, IEEE, 2020.

\bibitem{papoudakis2021agent}
G.~Papoudakis, F.~Christianos, and S.~Albrecht, ``Agent modelling under partial observability for deep reinforcement learning,'' {\em Advances in Neural Information Processing Systems}, vol.~34, pp.~19210--19222, 2021.

\bibitem{losey2020learning}
D.~P. Losey, M.~Li, J.~Bohg, and D.~Sadigh, ``Learning from my partner’s actions: Roles in decentralized robot teams,'' in {\em Conference on robot learning}, pp.~752--765, PMLR, 2020.

\bibitem{parekh2022rili}
S.~Parekh, S.~Habibian, and D.~P. Losey, ``Rili: Robustly influencing latent intent,'' in {\em 2022 IEEE/RSJ International Conference on Intelligent Robots and Systems (IROS)}, pp.~01--08, IEEE, 2022.

\bibitem{von2017minds}
F.~B. Von Der~Osten, M.~Kirley, and T.~Miller, ``The minds of many: Opponent modeling in a stochastic game.,'' in {\em IJCAI}, pp.~3845--3851, 2017.

\bibitem{ndousse2021emergent}
K.~K. Ndousse, D.~Eck, S.~Levine, and N.~Jaques, ``Emergent social learning via multi-agent reinforcement learning,'' in {\em International Conference on Machine Learning}, pp.~7991--8004, PMLR, 2021.

\bibitem{xie2021learning}
A.~Xie, D.~Losey, R.~Tolsma, C.~Finn, and D.~Sadigh, ``Learning latent representations to influence multi-agent interaction,'' in {\em Conference on robot learning}, pp.~575--588, PMLR, 2021.

\bibitem{wang2022influencing}
W.~Z. Wang, A.~Shih, A.~Xie, and D.~Sadigh, ``Influencing towards stable multi-agent interactions,'' in {\em Conference on robot learning}, pp.~1132--1143, PMLR, 2022.

\bibitem{jaques2019social}
N.~Jaques, A.~Lazaridou, E.~Hughes, C.~Gulcehre, P.~Ortega, D.~Strouse, J.~Z. Leibo, and N.~De~Freitas, ``Social influence as intrinsic motivation for multi-agent deep reinforcement learning,'' in {\em International conference on machine learning}, pp.~3040--3049, PMLR, 2019.

\bibitem{kim2022influencing}
D.-K. Kim, M.~Riemer, M.~Liu, J.~Foerster, M.~Everett, C.~Sun, G.~Tesauro, and J.~P. How, ``Influencing long-term behavior in multiagent reinforcement learning,'' {\em Advances in Neural Information Processing Systems}, vol.~35, pp.~18808--18821, 2022.

\bibitem{wright2015coordinate}
S.~J. Wright, ``Coordinate descent algorithms,'' {\em Mathematical programming}, vol.~151, no.~1, pp.~3--34, 2015.

\bibitem{le2022algames}
S.~Le~Cleac’h, M.~Schwager, and Z.~Manchester, ``Algames: a fast augmented lagrangian solver for constrained dynamic games,'' {\em Autonomous Robots}, vol.~46, no.~1, pp.~201--215, 2022.

\bibitem{scholler2020constant}
C.~Sch{\"o}ller, V.~Aravantinos, F.~Lay, and A.~Knoll, ``What the constant velocity model can teach us about pedestrian motion prediction,'' {\em IEEE Robotics and Automation Letters}, vol.~5, no.~2, pp.~1696--1703, 2020.

\bibitem{li2019cgns}
J.~Li, H.~Ma, and M.~Tomizuka, ``Conditional generative neural system for probabilistic trajectory prediction,'' pp.~6150--6156, 11 2019.

\bibitem{zhao2019multi}
T.~Zhao, Y.~Xu, M.~Monfort, W.~Choi, C.~Baker, Y.~Zhao, Y.~Wang, and Y.~N. Wu, ``Multi-agent tensor fusion for contextual trajectory prediction,'' in {\em Proceedings of the IEEE/CVF Conference on Computer Vision and Pattern Recognition}, pp.~12126--12134, 2019.

\bibitem{liang2019peeking}
J.~Liang, L.~Jiang, J.~C. Niebles, A.~G. Hauptmann, and L.~Fei-Fei, ``Peeking into the future: Predicting future person activities and locations in videos,'' in {\em Proceedings of the IEEE/CVF Conference on Computer Vision and Pattern Recognition}, pp.~5725--5734, 2019.

\bibitem{zhang2019sr}
P.~Zhang, W.~Ouyang, P.~Zhang, J.~Xue, and N.~Zheng, ``Sr-lstm: State refinement for lstm towards pedestrian trajectory prediction,'' in {\em Proceedings of the IEEE/CVF Conference on Computer Vision and Pattern Recognition}, pp.~12085--12094, 2019.

\bibitem{Huang2019stgat}
Y.~Huang, H.~Bi, Z.~Li, T.~Mao, and Z.~Wang, ``Stgat: Modeling spatial-temporal interactions for human trajectory prediction,'' in {\em 2019 IEEE/CVF International Conference on Computer Vision (ICCV)}, pp.~6271--6280, 2019.

\bibitem{Dendorfer_2020_ACCV}
P.~Dendorfer, A.~Osep, and L.~Leal-Taixe, ``Goal-gan: Multimodal trajectory prediction based on goal position estimation,'' in {\em Proceedings of the Asian Conference on Computer Vision (ACCV)}, November 2020.

\bibitem{shi2021sgcn}
L.~Shi, L.~Wang, C.~Long, S.~Zhou, M.~Zhou, Z.~Niu, and G.~Hua, ``Sgcn: Sparse graph convolution network for pedestrian trajectory prediction,'' in {\em Proceedings of the IEEE/CVF Conference on Computer Vision and Pattern Recognition}, pp.~8994--9003, 2021.

\bibitem{marchetti2020memnet}
F.~Marchetti, F.~Becattini, L.~Seidenari, and A.~Del~Bimbo, ``{MANTRA}: Memory augmented networks for multiple trajectory prediction,'' in {\em Proceedings of the IEEE Conference on Computer Vision and Pattern Recognition}, 2020.

\bibitem{shafiee2021introvert}
N.~Shafiee, T.~Padir, and E.~Elhamifar, ``Introvert: Human trajectory prediction via conditional 3d attention,'' in {\em Proceedings of the IEEE/CVF Conference on Computer Vision and Pattern Recognition}, pp.~16815--16825, 2021.

\bibitem{mangalam2020not}
K.~Mangalam, H.~Girase, S.~Agarwal, K.-H. Lee, E.~Adeli, J.~Malik, and A.~Gaidon, ``It is not the journey but the destination: Endpoint conditioned trajectory prediction,'' {\em arXiv preprint arXiv:2004.02025}, 2020.

\bibitem{pang2021trajectory}
B.~Pang, T.~Zhao, X.~Xie, and Y.~N. Wu, ``Trajectory prediction with latent belief energy-based model,'' in {\em Proceedings of the IEEE/CVF Conference on Computer Vision and Pattern Recognition}, pp.~11814--11824, 2021.

\bibitem{pecnet}
K.~Mangalam, H.~Girase, S.~Agarwal, K.-H. Lee, E.~Adeli, J.~Malik, and A.~Gaidon, ``It is not the journey but the destination: Endpoint conditioned trajectory prediction,'' in {\em Computer Vision--ECCV 2020: 16th European Conference, Glasgow, UK, August 23--28, 2020, Proceedings, Part II 16}, pp.~759--776, Springer, 2020.

\bibitem{amirian2020opentraj}
J.~Amirian, B.~Zhang, F.~V. Castro, J.~J. Baldelomar, J.-B. Hayet, and J.~Pettr{\'e}, ``Opentraj: Assessing prediction complexity in human trajectories datasets,'' in {\em Proceedings of the asian conference on computer vision}, 2020.

\bibitem{saadatnejad2023jrdb}
S.~Saadatnejad, Y.~Gao, H.~Rezatofighi, and A.~Alahi, ``Jrdb-traj: A dataset and benchmark for trajectory forecasting in crowds,'' {\em arXiv preprint arXiv:2311.02736}, 2023.

\bibitem{gao2022social}
Y.~Gao, ``Social-pose: Human trajectory prediction using input pose,'' 2022.

\bibitem{jrdb}
R.~Martin-Martin, M.~Patel, H.~Rezatofighi, A.~Shenoi, J.~Gwak, E.~Frankel, A.~Sadeghian, and S.~Savarese, ``Jrdb: A dataset and benchmark of egocentric robot visual perception of humans in built environments,'' {\em IEEE transactions on pattern analysis and machine intelligence}, vol.~45, no.~6, pp.~6748--6765, 2021.

\bibitem{yolo}
Y.~Zhang, Z.~Guo, J.~Wu, Y.~Tian, H.~Tang, and X.~Guo, ``Real-time vehicle detection based on improved yolo v5,'' {\em Sustainability}, vol.~14, no.~19, p.~12274, 2022.

\end{thebibliography}
\bibliographystyle{ieeetr}


\end{document}